# Face Recognition Methods & Applications

Divyarajsinh N. Parmar[1], Brijesh B. Mehta[2]
[1]P.G. Student of Computer Engineering
[2]Asst.Prof. Dept.of Computer Engineering
C.U. Shah College of Engg. & Tech
Wadhwan city, India
[1]parmar.raj1247@gmail.com
[2]brijeshbmehta@gmail.com

*Abstract* - Face recognition presents a challenging problem in the field of image analysis and computer vision. The security of information is becoming very significant and difficult. Security cameras are presently common in airports, Offices, University, ATM, Bank and in any locations with a security system. Face recognition is a biometric system used to identify or verify a person from a digital image. Face Recognition system is used in security. Face recognition system should be able to automatically detect a face in an image. This involves extracts its features and then recognize it, regardless of lighting, expression, illumination, ageing, transformations (translate, rotate and scale image) and pose, which is a difficult task. This paper contains three sections. The first section describes the common methods like holistic matching method, feature extraction method and hybrid methods. The second section describes applications with examples and finally third section describes the future research directions of face recognition.

## I. INTRODUCTION

Face recognition systems have been conducted now for almost 50 years. Face recognition is one of the researches in area pattern recognition & computer vision due to its numerous practical applications in the area of biometrics, Information security, access control, law enforcement, smart cards and surveillance system. The first large scale application of face recognition was carried out in Florida.

Biometric-based techniques have emerged as the most promising option for recognizing individuals in recent years since, instead of certifying people and allowing them access to physical and virtual domains based on passwords, PINs, smart cards, plastic cards, tokens, keys and so, these methods examine an individual's physiological and/or behavioral characteristics in order to determine and/or ascertain his/her identity. Passwords and PINs are difficult to remember and can be stolen or guessed; cards, tokens, keys and the like can be misplaced, forgotten, or duplicated; magnetic cards can become corrupted and unclear. However, an individual's biological traits cannot be misplaced, forgotten, stolen or forged [1].

In order to develop a useful and applicable face recognition system several factors need to be take in hand.
1. The overall speed of the system from detection to recognition should be acceptable.
2. The accuracy should be high
3. The system should be easily updated and enlarged, that is easy to increase the number of subjects that can be recognized.

## II. Face Recognition Methods

In the beginning of the 1970's, face recognition was treated as a 2D pattern recognition problem [2]. The distances between important points where used to recognize known faces, e.g. measuring the distance between the eyes or other important points or measuring different angles of facial components. But it is necessary that the face recognition systems to be fully automatic. Face recognition is such a challenging yet interesting problem that it has attracted researchers who have different backgrounds: psychology, pattern recognition, neural networks, computer vision, and computer graphics. The following methods are used to face recognition.

1. Holistic Matching Methods
2. Feature-based (structural) Methods
3. Hybrid Methods

**1. Holistic Matching Methods:** In holistic approach, the complete face region is taken into account as input data into face catching system. One of the best example of holistic methods are Eigenfaces [8] (most widely used method for face recognition), Principal Component Analysis, Linear Discriminant Analysis [7] and independent component analysis etc.

**1.1.1 Holistic example**
The first successful demonstration of machine recognition of faces was made by Turk and Pentland[2] in 1991 using eigenfaces. Their approach covers face recognition as a two-dimensional recognition problem. The flowchart in Figure 1 illustrates the different stages in an eigenface based recognition system. (1) The first stage is to insert a set of images into a database, these images are names as the training set and this is because they will be used when we compare images and when we create the eigenfaces. (2) The second stage is to create the eigenfaces. Eigenfaces are made by extracting characteristic features from the faces. The input images are normalized to line up the eyes and mouths. They are then resized so that they have the same size. Eigenfaces can now be extracted from the image data by using a mathematical tool called Principal Component Analysis (PCA). (3) When the eigenfaces have been created, each image will be represented as a vector of weights. (4) The system is now ready to accept entering queries. (5) The weight of the incoming unknown image is found and then compared to the weights of those already in the system. If the input image's weight is over a given threshold it is considered to be unidentified. The identification of the input image is done by finding the image in the database whose weights are the closest to the weights of the input image. The image in the database with the closest weight will be returned as a hit to the user of the system [2].

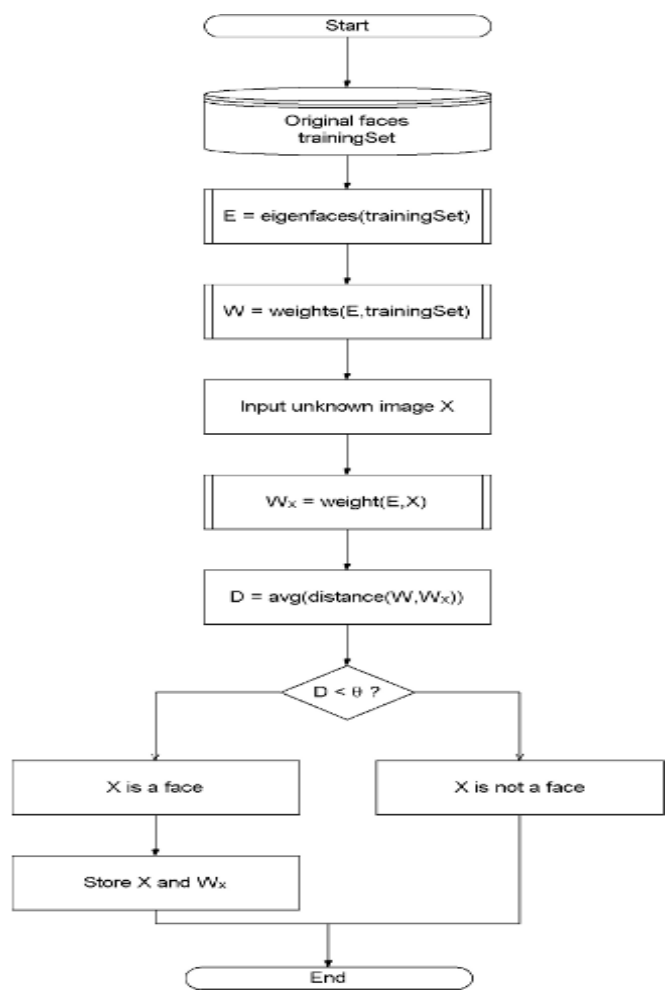

Figure 1: Flow chart of the eigenface-based algorithm.

**2. Feature-based (structural) Methods:** In this methods local features such as eyes, nose and mouth are first of all extracted and their locations and local statistics (geometric and/or appearance) are fed into a structural classifier. A big challenge for feature extraction methods is feature "restoration", this is when the system tries to retrieve features that are invisible due to large variations, e.g. head Pose when we are matching' a frontal image with a profile image.[5] Distinguishes between three different extraction methods:
  I. Generic methods based on edges, lines, and curves
  II. Feature-template-based methods
  III. Structural matching methods that take into consideration geometrical Constraints on the features.

**3. Hybrid Methods:** Hybrid face recognition systems use a combination of both holistic and feature extraction methods. Generally 3D Images are used in hybrid methods. The image of a person's face is caught in 3D, allowing the system to note the curves of the eye sockets, for example, or the shapes of the chin or forehead. Even a face in profile would serve because the system uses depth, and an axis of measurement, which gives it enough information to construct a full face. The 3D system usually proceeds thus: Detection, Position, Measurement, Representation and Matching. **Detection** - Capturing a face either a scanning a photograph or photographing a person's face in real time. **Position** - Determining the location, size and angle of the head. **Measurement** - Assigning measurements to each curve of the face to make a template with specific focus on the outside of the eye, the inside of the eye and the angle of the nose. **Representation** - Converting the template into a code - a numerical representation of the face and **Matching** - Comparing the received data with faces in the existing database.

In Case the 3D image is to be compared with an existing 3D image, it needs to have no alterations. Typically, however, photos that are put in 2D, and in that case, the 3D image need a few changes. This is tricky, and is one of the biggest challenges in the field today.

### III. Face Recognition Applications

Face recognition is also useful in human computer interaction, virtual reality, database recovery, multimedia, computer entertainment, information security e.g. operating system, medical records, online banking., Biometric e.g. Personal Identification - Passports, driver licenses , Automated identity verification - border controls , Law enforcement e.g. video surveillances , investigation , Personal Security - driver monitoring system, home video surveillance system.

**Applications & examples:**

**Face Identification:** Face recognition systems identify people by their face images. Face recognition systems establish the presence of an authorized person rather than just checking whether a valid identification (ID) or key is being used or whether the user knows the secret personal identification numbers (Pins) or passwords. The following are example.

To eliminate duplicates in a nationwide voter registration system because there are cases where the same person was assigned more than one identification number. The face recognition system directly compares the face images of the voters and does not use ID numbers to differentiate one from the others. When the top two matched faces are highly similar to the query face image, manual review is required to make sure they are indeed different persons so as to eliminate duplicates.

**Access Control:** In many of the access control applications, such as office access or computer logon, the size of the group of people that need to be recognized is relatively small. The face pictures are also caught under natural conditions, such as frontal faces and indoor illumination. The face recognition system of this application can achieve high accuracy without much co-operation from user. The following are the example. Face recognition technology is used to monitor continuously who is in front of a computer terminal. It allows the user to leave the terminal without closing files and logging out. When the user leaves for a predetermined time, a screen saver covers up the work and disables the mouse & keyboard. When the user comes back and is recognized, the screen saver clears and the previous session appears as it was left. Any other user who tries to logon without authorization is denied.

**Security:** Today more than ever, security is a primary concern at airports and for airline staff office and passengers. Airport protection systems that use face recognition technology have been implemented at many airports around the world. The following are the two examples.

In October, 2001, Fresno Yosemite International (FYI) airport in California deployed Viisage's face recognition technology for airport security purposes. The system is designed to alert FYl's airport public safety officers whenever an individual matching the appearance of a known terrorist suspect enters the airport's security checkpoint. Anyone recognized by the system would have further investigative processes by public safety officers. Computer security has also seen the application of face recognition technology. To prevent someone else from changing files or transacting with others when the authorized individual leaves the computer terminal for a short time, users are continuously authenticated, checking that the individual in front of the computer screen or at a user is the same authorized person who logged in.

**Image database investigations**: Searching image databases of licensed drivers, benefit recipients, missing children, immigrants and police bookings.

**General identity verification:** Electoral registration, banking, electronic commerce, identifying newborns, national IDs, passports, employee IDs.

**Surveillance:** Like security applications in public places, surveillance by face recognition systems has a low user satisfaction level, if not lower. Free lighting conditions, face orientations and other divisors all make the deployment of face recognition systems for large scale surveillance a challenging task. The following are some example of face-based surveillance.
To enhance town center surveillance in Newham Borough of London, this has 300 cameras linked to the closed circuit TV (CCTV) controller room. The city council claims that the technology has helped to achieve a 34% drop in crime since its facility. Similar systems are in place in Birmingham, England. In 1999 Visionics was awarded a contract from National Institute of Justice to develop smart CCTV technology.

IV. Conclusion & Scope for Future Research

It is our opinion that research in face recognition is an exciting area for many years to come and will keep many scientists and engineers busy. In this paper we have given concepts of face recognition methods & its applications. The present paper can provide the readers a better understanding about face recognition methods & applications.
In the future, 2D & 3D Face Recognition and large scale applications such as e-commerce, student ID, digital driver licenses, or even national ID is the challenging task in face recognition & the topic is open to further research.